# Handwritten Recognition Using SVM, KNN and Neural Network

Norhidayu binti Abdul Hamid
Advance Informatics School
Universiti Teknologi Malaysia
Kuala Lumpur, Malaysia
put3jaya22@gmail.com

Nilam Nur Binti Amir Sjarif*
Advance Informatics School
Universiti Teknologi Malaysia
Kuala Lumpur, Malaysia
nilamnur@utm.my

*Abstract*—Handwritten recognition (HWR) is the ability of a computer to receive and interpret intelligible handwritten input from source such as paper documents, photographs, touch-screens and other devices. In this paper we will using three (3) classification to recognize the handwritten which is SVM, KNN and Neural Network.

## I. INTRODUCTION

Handwriting recognition is the ability of a computer or device to take as input handwriting from source such as printed physical documents, pictures and other devices. It also can use handwriting as a direct input to a touch screen and then interpret it as text. There are many devices now can take handwriting as an input such as smartphones, tablets and PDA to a touch screen through a stylus or finger. This is useful as it allows the user to quickly write down number and text to the devices. There are many applications for handwriting recognition are available this day. There are many technique that have been developed to recognize the handwriting. One of them is Optical Character Recognition (OCR). OCR will read text from scanned document and translating the images into a form that computer can manipulate it. In this paper we will use three (3) classification algorithm to recognize the handwriting which is Support Vector Machine (SVM), K-Nearest Neighbor (KNN) and Neural Network. All this three (3) will be discuss later in this paper.

## II. RELATED WORK

There are many research have been done regarding the handwriting recognition in various field. Youssouf Chherawala, Partha Pratim Roy and Mohamed Cheriet in their paper "Feature Set Evaluation for Offline Handwriting Recognition Systems: Application to the Recurrent Neural Network" stated that handwriting recognition system is dependent on the features extracted from the word image. There are various method to extract the features but there are no method that have been proposed to identify the most promising of these other than a straightforward comparison based on the recognition rate. So they propose a framework for feature set evaluation based on a collaborative setting. They use a weighted vote combination of recurrent neural network (RNN) classifier. They quantify the importance of feature sets through the combination weights, which reflect their strength and complementarity.

Nurul Ilmi, Tjokorda Agung Budi W and Kurniawan Nur R in their paper "Handwriting Digit Recognition using Local Binary Pattern Variance and K-Nearest Neighbors Classification" using Local Binary Pattern (LBP) as feature extraction and K-NN classification on their handwriting recognition system on the C1 form used by General Elections Commission in Indonesia. The testing result is LBP variance can recognize handwriting digit character on MNIST dataset with accuracy 89.81% and for data from C1 form, the accuracy is 70.91%

In paper "Online Handwriting Verification with Safe Password and Increasing Number of Features", present a solution to verify user with safe handwritten password using Bayes Net, Naïve Bayes, K-NN and Multi-layer Perceptron (MLP) classifier. The main goal is to reduce the features to achieve the same or better result after ranking and reduce the processing time for classification. The result show that Bayes Net classifier is the best classifier with 100% correct for priority time, speed and relations and for FAR (False Acceptance Rate) of 3.13%.

Behnam Fallah and Hassan Khotanlou in their paper "Identify Human Personality parameters based on handwriting using neural network" suggest that the handwriting parameters anatomy will lead the psychologists to investigate the psychological principles of behavior, temperament, character, personality and the nervous and social aspects of a person's brain. They purpose a method to identify a person's personality through their handwriting. They applied Minnesota Multiphasic Personality Inventory to identify the character parameters. And they applied hidden Markov model and neural network (MLP) to perform classification to identify personality from handwriting. MLP was used to identify properties which not related to the writer and a hidden Markov model was used to classify those properties which are related to the writer. The advantages of the purpose algorithm as they claim is using dependent and independent features of text in the process of feature extraction, the proposed personality system in automated, particularly in the process of feature extraction, enhance accuracy and reliability of the personality recognition system due to the use of MMPI test on training step, no need to segmentation on feature extraction phase and using GDA to increase the space between classes.

### III. PROPOSED SYSTEM

For this system, we used python, openCV and sklearn to run classification and read the dataset. We used MNIST dataset for training and evaluation for classification. MNIST problem is a dataset for evaluating machine learning models on the handwritten digit classification problem. The dataset was constructed from a number of scanned document dataset available from the National Institute of Standards and Technology (NIST). Each image in this dataset is a 28 by 28 pixel square (748 pixels total). There are 70000 images in the dataset that can be used for training and evaluate the system.

In this proposed system we used three (3) classification algorithm to the recognition which is Support Vector Machine (SVM), K-Nearest Neighbor and Multi-layer Perceptron Neural Network (MLP)

#### A. Support Vector Machine (SVM)

SVM are a set of supervised learning methods used for classification, regression and outliers detection. In SVM, each data item will be plot as a point in n-dimensional space (n = number of feature) with the value of each feature being value of a particular coordinate. The classification is done by finding the hyper-plane that differentiate the two (2) classes.

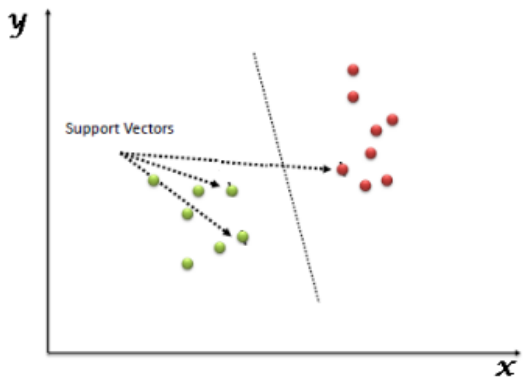

Figure 1 - SVM Classification

The SVM in scikit-learn support both dense (numpy.ndarray and convertible to that by numpy.asarray) and sparse (any scipy.sparse) sample vectors as input. In scikit-learn have three (3) classes that capable of performing multi-class classification on a dataset which is SVC, NuSVC and LinearSVC. In this system we will use LinearSVC class to perform the classification of MNIST dataset. LinearSVC or Linear Support Vector Classification that use linear kernel and implemented in terms of liblinear that has more flexibility in the choise of penalties and loss functions and should scale better large numbers of samples.

In this classification, we will have two (2) file which is generateClassifier.py for classification and performRecognition.py for testing the classification. In the generateClassifier.py we will perform three (3) step such as:

a. Calculate the HOG features for each sample in the database.

b. Train a multi-class linear SVM with the HOG features of each sample along with the corresponding label.

c. Save classifier in a file.

```
12 # Load the dataset
13 dataset = datasets.fetch_mldata("MNIST Original")
```

Figure 2 - Download MNIST Dataset

We will download the MNIST dataset as shown in figure 2. Then we will use Histogram of Oriented Gradients (HOG) feature detection to extract the feature of MNIST dataset. The dataset images of the digits will be save in a numpy array and corresponding labels. Next we will calculate the HOG features for each images and save them in another numpy array. The coding as shown in the figure 3 below.

```
15 # Extract the features and labels
16 features = np.array(dataset.data, 'int16')
17 labels = np.array(dataset.target, 'int')
18
19 # Extract the hog features
20 list_hog_fd = []
21 for feature in features:
22     fd = hog(feature.reshape((28, 28)), orientations=9, pixels_per_cell=(14, 14), cells_per
23     list_hog_fd.append(fd)
24 hog_features = np.array(list_hog_fd, 'float64')
```

Figure 3 - Features extraction

To calculate HOG features, we set the number of cell is of size 14 x 14. As we stated before MNIST dataset size is 28 x 28 pixel, so we will have four (4) blocks/cells of size 14 x 14 each. The orientation vector is set to 9. That mean HOG feature vector will be of size 4 x 9 = 36.

Then we will create Linear SVM object and perform the training for the dataset then we will save the classifier in a file as shown in figure 4 below.

```
32 # Create an linear SVM object
33 clf = LinearSVC()
34
35 # Perform the training
36 clf.fit(hog_features, labels)
37
38 # Save the classifier
39 joblib.dump((clf, pp), "digits_cls.pkl", compress=3)
```

Figure 4 - Linear SVM Classification

#### B. K-Nearest Neighbor (KNN)

KNN classifier is the most simple image classification algorithm. KNN classification doesn't actually learn anything. This algorithm is relies on the distance between feature vectors. KNN algorithm classifies unknown data points by finding the most common class among the k closest examples. Each data point in the k closest cast a vote and the highest category number of votes wins.

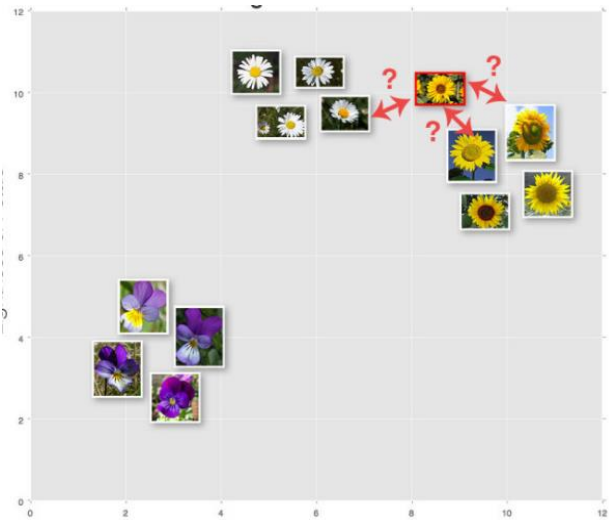
Figure 5 - K-NN Classification

There are five (5) step to train image classifier which is:
a. Step 1 – Structuring our initial dataset
b. Step 2 – Splitting the dataset
c. Step 3 - Extracting features
d. Step 4 - Training the classification model
e. Step 5 – Evaluating classifier

First we will load the MNIST dataset into the system. Then we will divide the dataset into training and testing data. We will divide the dataset into 75% for training and 25% for testing. From the training dataset we will spare 10% for validation as shown in figure 6 below.

```
12  # load the MNIST digits dataset
13  mnist = datasets.load_digits()
14
15  # take the MNIST data and construct the training and testing split, using 75% of the
16  # data for training and 25% for testing
17  (trainData, testData, trainLabels, testLabels) = train_test_split(np.array(mnist.data),
18      mnist.target, test_size=0.25, random_state=42)
19
20  # now, let's take 10% of the training data and use that for validation
21  (trainData, valData, trainLabels, valLabels) = train_test_split(trainData, trainLabels,
22      test_size=0.1, random_state=84)
```
Figure 6 - Split training, testing and validation

After that we will train the classifier and find the optimal value of k. we also will calculate the accuracy of the classifier. In this classifier, we will loop k value between 1 to 15. Then we will validate the classifier as shown in figure 7 below.

```
34  # loop over various values of `k` for the k-Nearest Neighbor classifier
35  for k in range(1, 30, 2):
36      # train the k-Nearest Neighbor classifier with the current value of `k`
37      model = KNeighborsClassifier(n_neighbors=k)
38      model.fit(trainData, trainLabels)
39
40      # evaluate the model and update the accuracies list
41      score = model.score(valData, valLabels)
42      print("k=%d, accuracy=%.2f%%" % (k, score * 100))
43      accuracies.append(score)
44
45  # find the value of k that has the largest accuracy
46  i = np.argmax(accuracies)
47  print("k=%d achieved highest accuracy of %.2f%% on validation data" % (kVals[i],
48      accuracies[i] * 100))
```
Figure 7 - Training and Validation Classifier

Then we will use classifier on the testing data. We will use k value equal to 1. Then we will perform the final evaluation as shown in figure 7.

```
50  # re-train our classifier using the best k value and predict the labels of the
51  # test data
52  model = KNeighborsClassifier(n_neighbors=kVals[i])
53  model.fit(trainData, trainLabels)
54  predictions = model.predict(testData)
55
56  # show a final classification report demonstrating the accuracy of the classifier
57  # for each of the digits
58  print("EVALUATION ON TESTING DATA")
59  print(classification_report(testLabels, predictions))
```
Figure 8 - Testing classifier

Then we will examining some individual predictions. We will loop over five (5) random images from testing data set.

```
61  # loop over a few random digits
62  for i in np.random.randint(0, high=len(testLabels), size=(5,)):
63      # grab the image and classify it
64      image = testData[i]
65      prediction = model.predict(image)[0]
66
67      # convert the image for a 64-dim array to an 8 x 8 image compatible with OpenCV,
68      # then resize it to 32 x 32 pixels so we can see it better
69      image = image.reshape((8, 8)).astype("uint8")
70      image = exposure.rescale_intensity(image, out_range=(0, 255))
71      image = imutils.resize(image, width=32, inter=cv2.INTER_CUBIC)
72
73      # show the prediction
74      print("I think that digit is: {}".format(prediction))
75      cv2.imshow("Image", image)
76      cv2.waitKey(0)
```
Figure 9 - Examining the classifier

C. *Multi-layer Perceptron Neural Network (MLP)*

Multi-layer Perceptron (MLP) is a supervised algorithm that learns a function:

$$f(\cdot) : R^m \rightarrow R^o$$

The function will learn by training on a dataset, where *m* is the number of dimensions for input and *o* is the number of dimensions for output. Given a set of features X = x1, x2, ….., xm and a target *y*, it can learn a non-linear function approximator for either classification or regression. It is different from logistic regression, in that between the input and the output layer, there can be one or more non-linear layers, call hidden layer.

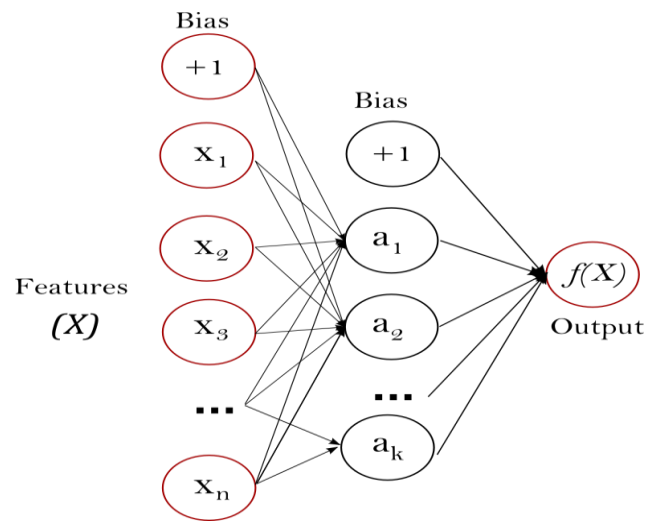
Figure 10 - MLP hidden layer

In this classification, we will have two (2) file which is generateClassifier-nn.py for classification and performRecognition.py for testing the classification. In the generateClassifier.py we will perform three (3) step such as:

a. Calculate the HOG features for each sample in the database.
b. Train a multi-class MLP neural network with the HOG features of each sample along with the corresponding label.
c. Save classifier in a file.

```
12 # Load the dataset
13 dataset = datasets.fetch_mldata("MNIST Original")
```

Figure 11 - Download MNIST Dataset

We will download the MNIST dataset as shown in figure 2. Then we will use Histogram of Oriented Gradients (HOG) feature detection to extract the feature of MNIST dataset. The dataset images of the digits will be save in a numpy array and corresponding labels. Next we will calculate the HOG features for each images and save them in another numpy array. The coding as shown in the figure 12 below.

```
15 # Extract the features and labels
16 features = np.array(dataset.data, 'int16')
17 labels = np.array(dataset.target, 'int')
18
19 # Extract the hog features
20 list_hog_fd = []
21 for feature in features:
22     fd = hog(feature.reshape((28, 28)), orientations=9, pixels_per_cell=(14, 14), cells_per
23     list_hog_fd.append(fd)
24 hog_features = np.array(list_hog_fd, 'float64')
```

Figure 12 - Features extraction

To calculate HOG features, we set the number of cell is of size 14 x 14. As we stated before MNIST dataset size is 28 x 28 pixel, so we will have four (4) blocks/cells of size 14 x 14 each. The orientation vector is set to 9. That mean HOG feature vector will be of size 4 x 9 = 36.

Then we will create MLP Neural Network object and perform the training for the dataset then we will save the classifier in a file as shown in figure 13 below.

```
29
30 # Create an MLP Neural Network object
31 clf = MLPClassifier(solver='lbfgs', alpha=1e-5,
32                    hidden_layer_sizes=(5, 2), random_state=1)
33
34 # Perform the training
35 clf.fit(hog_features, labels)
36
```

Figure 13 – MLP Neural Network Classification

D. *Experiment*

For SVM and MLP Neural Network we will test the classifier using performRecognition.py. In this script, we will classifier file that we save before as shown in figure 14.

```
16 # Load the classifier
17 clf, pp = joblib.load(args["classiferPath"])
18
```

Figure 14 - Call classifier file

Then we will load the image that will be used to test the classifier. Then we will preprocessing the image like convert to grayscale and apply Gaussian filtering. Then we will threshold the image and find contours in the image. Then we will get the rectangles contains each contour as shown in figure 15.

```
19 # Read the input image
20 im = cv2.imread(args["image"])
21
22 # Convert to grayscale and apply Gaussian filtering
23 im_gray = cv2.cvtColor(im, cv2.COLOR_BGR2GRAY)
24 im_gray = cv2.GaussianBlur(im_gray, (5, 5), 0)
25
26 # Threshold the image
27 ret, im_th = cv2.threshold(im_gray, 90, 255, cv2.THRESH_BINARY_INV)
28
29 # Find contours in the image
30 ctrs, hier = cv2.findContours(im_th.copy(), cv2.RETR_EXTERNAL, cv2.CHAIN_APPROX_SIMPLE)
31
32 # Get rectangles contains each contour
33 rects = [cv2.boundingRect(ctr) for ctr in ctrs]
```

Figure 15 - Preprocessing the image

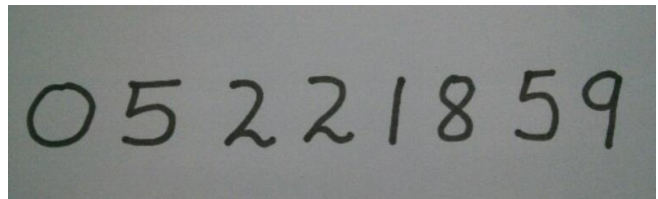

Figure 16 - Testing image

Then we will perform the classifier to the testing image. We will calculate the HOG features and predict the digit of the image and then we will show the result image as shown in figure 17.

```
35 # For each rectangular region, calculate HOG features and predict
36 # the digit using classifier
37 for rect in rects:
38     # Draw the rectangles
39     cv2.rectangle(im, (rect[0], rect[1]), (rect[0] + rect[2], rect[1] + rect[3]), (0, 255,
40     # Make the rectangular region around the digit
41     leng = int(rect[3] * 1.6)
42     pt1 = int(rect[1] + rect[3] // 2 - leng // 2)
43     pt2 = int(rect[0] + rect[2] // 2 - leng // 2)
44     roi = im_th[pt1:pt1+leng, pt2:pt2+leng]
45     # Resize the image
46     roi = cv2.resize(roi, (28, 28), interpolation=cv2.INTER_AREA)
47     roi = cv2.dilate(roi, (3, 3))
48     # Calculate the HOG features
49     roi_hog_fd = hog(roi, orientations=9, pixels_per_cell=(14, 14), cells_per_block=(1, 1),
50     roi_hog_fd = pp.transform(np.array([roi_hog_fd], 'float64'))
51     nbr = clf.predict(roi_hog_fd)
52     cv2.putText(im, str(int(nbr[0])), (rect[0], rect[1]),cv2.FONT_HERSHEY_DUPLEX, 2, (0, 25
53
54 cv2.namedWindow("Resulting Image with Rectangular ROIs", cv2.WINDOW_NORMAL)
55 cv2.imshow("Resulting Image with Rectangular ROIs", im)
56 cv2.waitKey()
57
```

Figure 17 - Examining the classifier

For KNN, we will perform the experiment as describe in the previous segment in KNN that we will examining the classifier using random images from testing data set.

IV. RESULT

For KNN, the result shows for value k equal 1 to 15 have the same result which is 99.26% which the highest accuracy as shown in figure 18.

Figure 18 - Accuracy of the classifier

The evaluation on testing data result as shown in figure 19 below.

Figure 19 - Evaluation on testing data

For the examining classifier with the five (5) random image from training dataset as shown in figure 20.

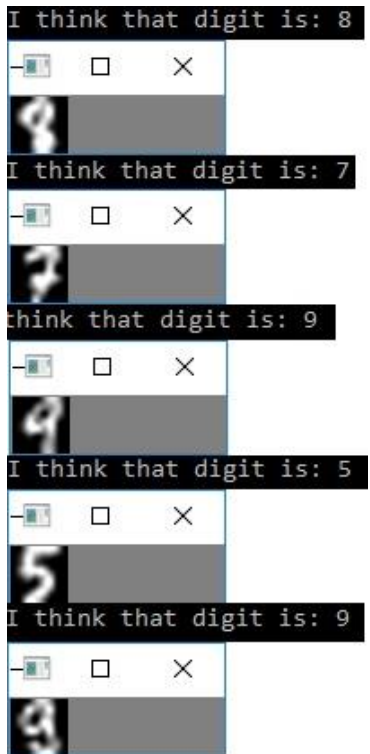

Figure 20 - Examining result

For the result of SVM classification is shown in figure 21 below.

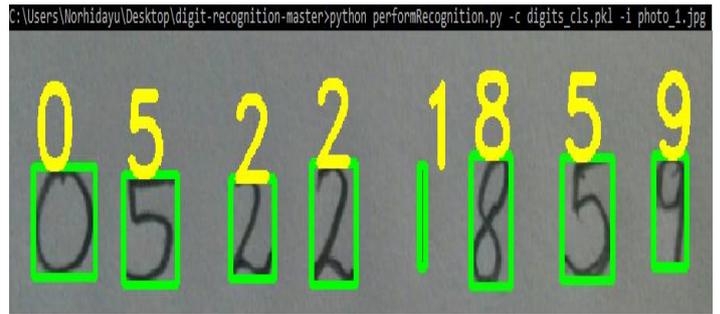

Figure 21 - SVM Classifier result

For the result of MLP Neural Network classification is shown in figure 22 below.

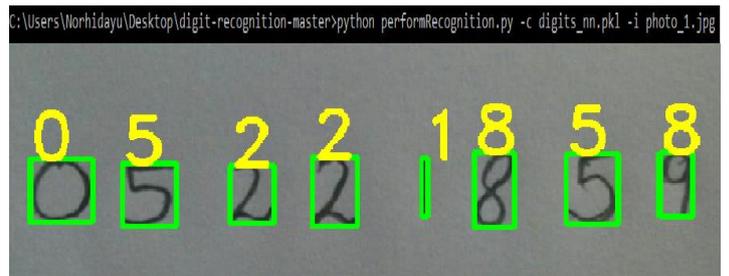

Figure 22 - MLP Neural Network Result

## V. CONCLUSION

As we can see from the result, KNN and SVM predict correctly the dataset but for MLP Neural Network that is some mistake to predict the number 9. This is because for KNN and SVM it predict directly from the feature extraction. But for MLP, it is a non-linear function. So it more suitable for learn non-linear models. And MLP with hidden layers have non-convex loss function where there exists more than one local minimum. Therefore different random weight initializations can lead to different validation accuracy. But it can improve by using Convolutional Neural Networks with Keras.

**generateClassifier.py**

```python
#!/usr/bin/python

# Import the modules
from sklearn.externals import joblib
from sklearn import datasets
from skimage.feature import hog
from sklearn.svm import LinearSVC
from sklearn import preprocessing
import numpy as np
from collections import Counter

# Load the dataset
dataset = datasets.fetch_mldata("MNIST Original")

# Extract the features and labels
features = np.array(dataset.data, 'int16')
labels = np.array(dataset.target, 'int')

# Extract the hog features
list_hog_fd = []
for feature in features:
    fd = hog(feature.reshape((28, 28)), orientations=9, pixels_per_cell=(14, 14), cells_per_block=(1, 1), visualise=False)
    list_hog_fd.append(fd)
hog_features = np.array(list_hog_fd, 'float64')

# Normalize the features
pp = preprocessing.StandardScaler().fit(hog_features)
hog_features = pp.transform(hog_features)

print ("Count of digits in dataset", Counter(labels))

# Create an linear SVM object
clf = LinearSVC()

# Perform the training
clf.fit(hog_features, labels)

# Save the classifier
joblib.dump((clf, pp), "digits_cls.pkl", compress=3)
```

**performClassifier.py**

```python
#!/usr/bin/python

# Import the modules
import cv2
from sklearn.externals import joblib
from skimage.feature import hog
import numpy as np
import argparse as ap

# Get the path of the training set
parser = ap.ArgumentParser()
parser.add_argument("-c", "--classiferPath", help="Path to Classifier File", required="True")
parser.add_argument("-i", "--image", help="Path to Image", required="True")
```

```python
args = vars(parser.parse_args())

# Load the classifier
clf, pp = joblib.load(args["classiferPath"])

# Read the input image
im = cv2.imread(args["image"])

# Convert to grayscale and apply Gaussian filtering
im_gray = cv2.cvtColor(im, cv2.COLOR_BGR2GRAY)
im_gray = cv2.GaussianBlur(im_gray, (5, 5), 0)

# Threshold the image
ret, im_th = cv2.threshold(im_gray, 90, 255, cv2.THRESH_BINARY_INV)

# Find contours in the image
ctrs, hier = cv2.findContours(im_th.copy(), cv2.RETR_EXTERNAL, cv2.CHAIN_APPROX_SIMPLE)

# Get rectangles contains each contour
rects = [cv2.boundingRect(ctr) for ctr in ctrs]

# For each rectangular region, calculate HOG features and predict
# the digit using classifier.
for rect in rects:
    # Draw the rectangles
    cv2.rectangle(im, (rect[0], rect[1]), (rect[0] + rect[2], rect[1] + rect[3]), (0, 255, 0), 3)
    # Make the rectangular region around the digit
    leng = int(rect[3] * 1.6)
    pt1 = int(rect[1] + rect[3] // 2 - leng // 2)
    pt2 = int(rect[0] + rect[2] // 2 - leng // 2)
    roi = im_th[pt1:pt1+leng, pt2:pt2+leng]
    # Resize the image
    roi = cv2.resize(roi, (28, 28), interpolation=cv2.INTER_AREA)
    roi = cv2.dilate(roi, (3, 3))
    # Calculate the HOG features
    roi_hog_fd = hog(roi, orientations=9, pixels_per_cell=(14, 14), cells_per_block=(1, 1), visualise=False)
    roi_hog_fd = pp.transform(np.array([roi_hog_fd], 'float64'))
    nbr = clf.predict(roi_hog_fd)
    cv2.putText(im, str(int(nbr[0])), (rect[0], rect[1]),cv2.FONT_HERSHEY_DUPLEX, 2, (0, 255, 255), 3)

cv2.namedWindow("Resulting Image with Rectangular ROIs", cv2.WINDOW_NORMAL)
cv2.imshow("Resulting Image with Rectangular ROIs", im)
cv2.waitKey()
```

**generateClassifier-nn.py**

```python
# Import the modules
from sklearn.externals import joblib
from sklearn import datasets
from skimage.feature import hog
from sklearn.neural_network import MLPClassifier
from sklearn import preprocessing
import numpy as np
from collections import Counter

# Load the dataset
dataset = datasets.fetch_mldata("MNIST Original")
```

```python
# Extract the features and labels
features = np.array(dataset.data, 'int16')
labels = np.array(dataset.target, 'int')

# Extract the hog features
list_hog_fd = []
for feature in features:
    fd = hog(feature.reshape((28, 28)), orientations=9, pixels_per_cell=(14, 14), cells_per_block=(1, 1), visualise=False)
    list_hog_fd.append(fd)
hog_features = np.array(list_hog_fd, 'float64')

# Normalize the features
pp = preprocessing.StandardScaler().fit(hog_features)
hog_features = pp.transform(hog_features)

print ("Count of digits in dataset", Counter(labels))

# Create an MLP Neural Network object
clf = MLPClassifier(solver='lbfgs', alpha=1e-5,
            hidden_layer_sizes=(5, 2), random_state=1)

# Perform the training
clf.fit(hog_features, labels)

# Save the classifier
joblib.dump((clf, pp), "digits_nn.pkl", compress=3)
```

**knn.py**

```python
# import the necessary packages
from __future__ import print_function
from sklearn.cross_validation import train_test_split
from sklearn.neighbors import KNeighborsClassifier
from sklearn.metrics import classification_report
from sklearn import datasets
from skimage import exposure
import numpy as np
import imutils
import cv2

# load the MNIST digits dataset
mnist = datasets.load_digits()

# take the MNIST data and construct the training and testing split, using 75% of the
# data for training and 25% for testing
(trainData, testData, trainLabels, testLabels) = train_test_split(np.array(mnist.data),
        mnist.target, test_size=0.25, random_state=42)

# now, let's take 10% of the training data and use that for validation
(trainData, valData, trainLabels, valLabels) = train_test_split(trainData, trainLabels,
        test_size=0.1, random_state=84)

# show the sizes of each data split
print("training data points: {}".format(len(trainLabels)))
print("validation data points: {}".format(len(valLabels)))
print("testing data points: {}".format(len(testLabels)))
```

```python
# initialize the values of k for our k-Nearest Neighbor classifier along with the
# list of accuracies for each value of k
kVals = range(1, 30, 2)
accuracies = []

# loop over various values of `k` for the k-Nearest Neighbor classifier
for k in range(1, 30, 2):
    # train the k-Nearest Neighbor classifier with the current value of `k`
    model = KNeighborsClassifier(n_neighbors=k)
    model.fit(trainData, trainLabels)

    # evaluate the model and update the accuracies list
    score = model.score(valData, valLabels)
    print("k=%d, accuracy=%.2f%%" % (k, score * 100))
    accuracies.append(score)

# find the value of k that has the largest accuracy
i = np.argmax(accuracies)
print("k=%d achieved highest accuracy of %.2f%% on validation data" % (kVals[i],
    accuracies[i] * 100))

# re-train our classifier using the best k value and predict the labels of the
# test data
model = KNeighborsClassifier(n_neighbors=kVals[i])
model.fit(trainData, trainLabels)
predictions = model.predict(testData)

# show a final classification report demonstrating the accuracy of the classifier
# for each of the digits
print("EVALUATION ON TESTING DATA")
print(classification_report(testLabels, predictions))

# loop over a few random digits
for i in np.random.randint(0, high=len(testLabels), size=(5,)):
    # grab the image and classify it
    image = testData[i]
    prediction = model.predict(image)[0]

    # convert the image for a 64-dim array to an 8 x 8 image compatible with OpenCV,
    # then resize it to 32 x 32 pixels so we can see it better
    image = image.reshape((8, 8)).astype("uint8")
    image = exposure.rescale_intensity(image, out_range=(0, 255))
    image = imutils.resize(image, width=32, inter=cv2.INTER_CUBIC)

    # show the prediction
    print("I think that digit is: {}".format(prediction))
    cv2.imshow("Image", image)
    cv2.waitKey(0)
```